\documentclass[10pt,twocolumn,letterpaper]{article}

\usepackage[wacv]{wacv}      % To produce the REVIEW version for the algorithms track
\usepackage{graphicx}
\usepackage{amsmath}
\usepackage{amssymb}
\usepackage{booktabs}
\usepackage{float}
\usepackage{enumitem}
\usepackage[table]{xcolor} % Include xcolor package with table option
\definecolor{lightred}{RGB}{255,200,200}
\definecolor{lightyellow}{RGB}{255,255,200}
\usepackage{soul}

\usepackage[pagebackref,breaklinks,colorlinks]{hyperref}

\usepackage[capitalize]{cleveref}
\crefname{section}{Sec.}{Secs.}
\Crefname{section}{Section}{Sections}
\Crefname{table}{Table}{Tables}
\crefname{table}{Tab.}{Tabs.}

\def\wacvPaperID{2334} % *** Enter the WACV Paper ID here
\def\confName{WACV}
\def\confYear{2025}

\begin{document}

\title{UW-GS: Distractor-Aware 3D Gaussian Splatting for Enhanced Underwater Scene Reconstruction}

\author{%
  Haoran Wang,
  Nantheera Anantrasirichai,
  Fan Zhang, and 
  David Bull \\
  School of Computer Science,
  University of Bristol,
  Bristol, UK \\\tt\small{\{yp22378,n.anantrasirichai,fan.zhang,dave.bull\}@bristol.ac.uk}}
\maketitle

%%%%%%%%% ABSTRACT
\begin{abstract}
3D Gaussian splatting (3DGS) offers the capability to achieve real-time high quality 3D scene rendering. However, 3DGS assumes that the scene is in a clear medium environment and struggles to generate satisfactory representations in underwater scenes, where light absorption and scattering are prevalent and moving objects are involved. To overcome these, we introduce a novel Gaussian Splatting-based method, UW-GS, designed specifically for underwater applications. It introduces a color appearance that models distance-dependent color variation, employs a new physics-based density control strategy to enhance clarity for distant objects, and uses a binary motion mask to handle dynamic content. Optimized with a well-designed loss function supporting for scattering media and strengthened by pseudo-depth maps, UW-GS outperforms existing methods with PSNR gains up to 1.26dB. To fully verify the effectiveness of the model, we also developed a new underwater dataset, S-UW, with dynamic object masks. The code of UW-GS and S-UW will be available at \url{https://github.com/WangHaoran16/UW-GS}.%Demonstrating its excellent performance, all while maintaining relatively high efficiency. achieves an advancement in underwater 3D scene modeling. Additionally, we provide a new underwater dataset with dynamic object masking.
\end{abstract}
\begin{figure*}[!t]
  \centering

   \includegraphics[width=1.0\linewidth]{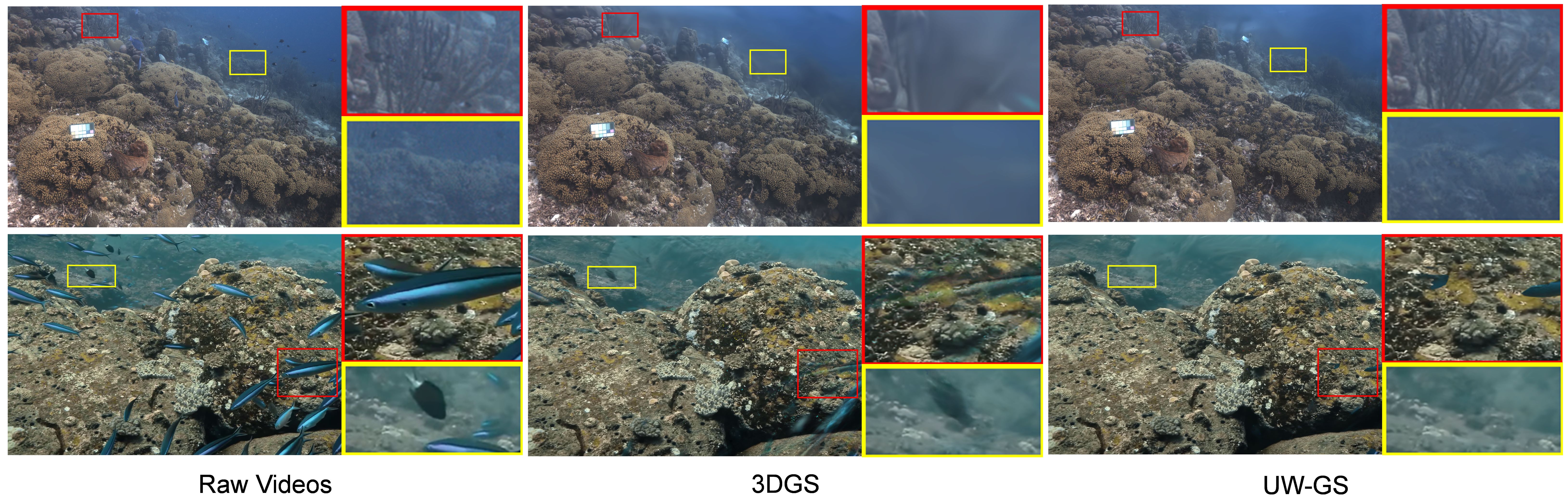}

   \caption{Visual comparison between 3DGS \cite{kerbl20233d} and our proposed UW-GS method. Left to right: Raw videos and the results of 3DGS and UW-GS, respectively. The top row, enhanced for visualization purposes, shows that UW-GS produces sharper results than 3DGS. The bottom row demonstrates UW-GS's better capability in handling moving objects.
   }
   \label{fig:demo}
\end{figure*}

%%%%%%%%% BODY TEXT
\section{Introduction}
\label{sec:intro}
% Motivation and importance
Ocean exploration has gained increasing importance in recent years, driven by the applications in underwater archaeology, geology, and other marine sciences. However, underwater activities are often constrained by the limitations of current technologies, the scarcity of diving experts, and high operational costs. As a result, 3D scene reconstruction has become crucial, as it allows underwater data to be captured, visualized, and analyzed onshore, facilitating more extensive and detailed studies without the need for continuous physical presence underwater.

% current methods 
The latest advances in 3D scene representations typically employ Neural Radiance Field (NeRF) \cite{mildenhall2020nerf} and 3D Gaussian Splatting (3DGS) \cite{kerbl20233d} based methods to model static video content captured in the wild. NeRF models implicitly represent a scene as a radiance field parameterized by a deep neural network that predicts volume density and colors. Although it offers effective view synthesis, training NeRF requires substantial computational time. Moreover, the implicit neural radiance fields also tend to yield suboptimal reconstruction results \cite{gao2024mani}. In contrast, 3DGS provides a faster alternative for training and rendering \cite{chen2024survey}, which represents a scene using a set of points in space, referred to as ``Gaussians", and generally delivers better quality.

% resarch gap
Despite the rapid growth of NeRFs and 3DGSs, most methods are designed for clear media, while only a few studies have attempted to reconstruct 3D scenes in diffuse environments, such as haze and underwater \cite{ramazzina2023scatternerf}. Underwater scenes, in particular, face highly variable lighting conditions influenced by factors such as water depth, surface waves, and water clarity. Visual content in these settings often experiences significant attenuation and degradation due to \textbf{\textit{light absorption}} and \textbf{\textit{backscattering}}, leading to inaccuracies in color and density estimations in the 3D reconstruction methods, especially 3DGS, as shown in the upper row \autoref{fig:demo}. Additionally, underwater scenes typically contain \textbf{\textit{moving elements}}, such as fish and floating debris, increasing complexity. These elements are frequently referred to as distractors in 3D modeling because they are not consistently present and create artifacts in some viewpoints \cite{sabour2023robustnerf}, as seen in ~\autoref{fig:demo} bottom row. To deal with them, the models must accurately differentiate between static and dynamic components, a task that becomes more challenging in a fluid medium. Unfortunately, the existing methods \cite{levy2023seathru, tang2024neural} do not address this issue.

% brief proposed method
To address the aforementioned issues, we propose a new Gaussian Splatting (GS)-based method, UW-GS, specifically for underwater scenes. Unlike traditional 3DGS methods, which struggle with light scattering due to their use of spherical harmonics, our approach includes a novel color appearance model that integrates medium effects into Gaussian attributes and employs affine transformations to account for appearance variations at different distances. We also introduce a physics-based density control strategy to enhance the clarity of distant objects by ensuring an adequate Gaussian representation. To manage distractors, we have developed a robustness estimation algorithm for GS that utilizes a binary motion mask during rendering to minimize the impact of dynamic content on training. The proposed UW-GS is optimized using a new loss function based on the characteristics of scattering media, to reduce blurring effects and enable more accurate medium effect estimation. We also incorporated pseudo-depth maps generated from DepthAnything \cite{yang2024depth}, trained with more general scenes, to enhance the robustness of our method. Finally, given the scarcity of underwater datasets, we collected a new dataset featuring four expansive areas of shallow underwater scenes, each presenting unique challenges compared to existing datasets. We also provide dynamic object masking maps in this database to facilitate further study and evaluation. In summary, our key contributions are as follows:

\begin{enumerate}[label=\arabic*), leftmargin=13pt, nolistsep]
\item We, for the first time, propose a Gaussian Splatting based solution for underwater scene representation with distractor awareness, UW-GS. % This proposed method proves robust across a wide range of underwater conditions.
\item UW-GS employs a novel color appearance model, transforming Gaussian colors to address low contrast and color distortion issues due to light attenuation and scattering.%y to solve the densification failure issues caused by heavy light absorption and scattering in the far regions.
\item We propose a new physics-based density control strategy aimed at preventing sparse Gaussians in highly absorptive and scattering media. This approach significantly enhances the clarity of details in distant areas.%, as shown in the first row of \autoref{fig:demo}.
\item We propose a new loss function that combines scattering medium characteristics with pseudo-depth maps and a pixel-wise binary motion mask to address estimated depth ambiguity and moving distractors in underwater scenes, respectively. %Please refer to the second row of \autoref{fig:demo}.
\item We introduce the first shallow underwater dataset, complete with associated distractor markers. This dataset presents unique challenges, including light flickering and water currents, adding complexity to 3D reconstruction.
%with binary masks for dynamic objects.
\end{enumerate}

%\hl{[HW: We need to include a short summary of the results. for example, we tested our approach on xxx datasets, what performance achieved (in PSNR), and ref figure 1 for visual comparison.]}

%------------------------------------------------------------------------
\section{Related Work}
\label{sec:relatedworks}

%This section briefly reviews underwater computer vision development and gives an overview of the 3D scene representation method, including NeRF \cite{ wang2021nerf } and 3DGS \cite{kerbl20233d}.
%-------------------------------------------------------------------------
\noindent\textbf{Underwater Imaging.} Underwater environments often suffer from issues such as color cast, image distortion, and contrast loss, largely due to the physical properties of water and its interaction with light \cite{gonzalez2023survey}. Given the similarities between haze and underwater image processing, techniques such as the dark-channel prior \cite{he2010single}, gray-world assumption \cite{ebner2009color}, and histogram equalization \cite{gonzalez2009digital} have been employed. Moreover, several studies have modeled underwater light behaviors, including the characterization of underwater light propagation \cite{dekker2001imaging}, radiative transfer theory \cite{dutre2018advanced}, and light reflectance models \cite{dekker2001imaging}. The Revised Underwater Image Formation Model (UIFM) \cite{akkaynak2018revised} is currently the most widely used, noting that attenuation and backscattering factors for direct and backscattered signals differ and are thus modeled independently. Various works in underwater image processing \cite{kar2021zero, huo2021efficient, zhang2024atlantis} have achieved remarkable results using this model. However, relying solely on single-image processing can yield inconsistent results across similar scenarios, limiting the reliability and effectiveness of these approaches, particularly used for 3D modeling.

%To enhance our understanding of real-world underwater environments, we will also conduct the knowledge of underwater imaging to the 3d scene representation method.

%-------------------------------------------------------------------------
\noindent\textbf{Neural Radiance Fields (NeRFs)} \cite{mildenhall2020nerf} have significantly advanced the field of 3D reconstruction in recent years. NeRF utilizes a radiance field—a volumetric representation that calculates density and color for each sampled point along a ray \cite{croce2023neural}. Subsequent NeRF-based techniques have improved synthesis through various methods, including cone tracing instead of traditional ray tracing \cite{barron2021mip, barron2022mip}, super-resolution techniques \cite{li2023uhdnerf, huang2023refsr}, grid-based methods \cite{barron2023zip, wang2023f2, tang2023delicate}, and deblurring \cite{wang2023bad}. Moreover, more compact models using fewer parameters have been developed \cite{xie2023hollownerf}, and faster rendering techniques have been introduced \cite{li2023nerfacc}. NeRF has also been extended to dynamic \cite{li2021neural} and few-shot rendering scenarios \cite{somraj2023vip}. However, these methods generally struggle in scattering media. To address this, scatterNeRF \cite{ramazzina2023scatternerf} was developed, effectively distinguishing objects from the medium in foggy scenes. SeaThru-NeRF \cite{levy2023seathru} utilizes a novel architecture to determine water medium coefficients based on the revised underwater image formation model \cite{akkaynak2018revised}. Moreover, other work \cite{tang2024neural} aims to accurately represent underwater scenes, incorporating dynamic rendering and tone mapping to handle variable illumination. Despite these advancements, NeRF-based methods still face challenges with the slow training and rendering times, especially in large-sized scenes.

\noindent\textbf{3D Gaussian Splatting (3DGS)}\cite{kerbl20233d} is an explicit representation method that utilizes a set of trainable 3D Gaussian points. Due to its fast training speed and high-quality real-time rendering capabilities, 3DGS has become a highly popular technology in computer graphics. Leveraging these advantages, our method is also developed based on 3DGS. Numerous enhancements have been proposed to optimize this approach for challenging data. These improvements include techniques for deblurring \cite{chen2024deblur, lee2024deblurring}, anti-aliasing \cite{yu2024mip, yan2024multi}, frequency regularization \cite{zhang2024fregs}, and refined density control strategies \cite{zhang2024pixel}. Additionally, 3DGS has been adapted for 4D rendering \cite{Wu_2024_CVPR} to accommodate dynamic scene rendering. Despite these advances, the inherent properties of 3DGS still pose limitations on its performance in complex underwater environments \cite{li2024watersplatting}.

\begin{figure*}[!t]
  \centering

   \includegraphics[width=1.0\linewidth]{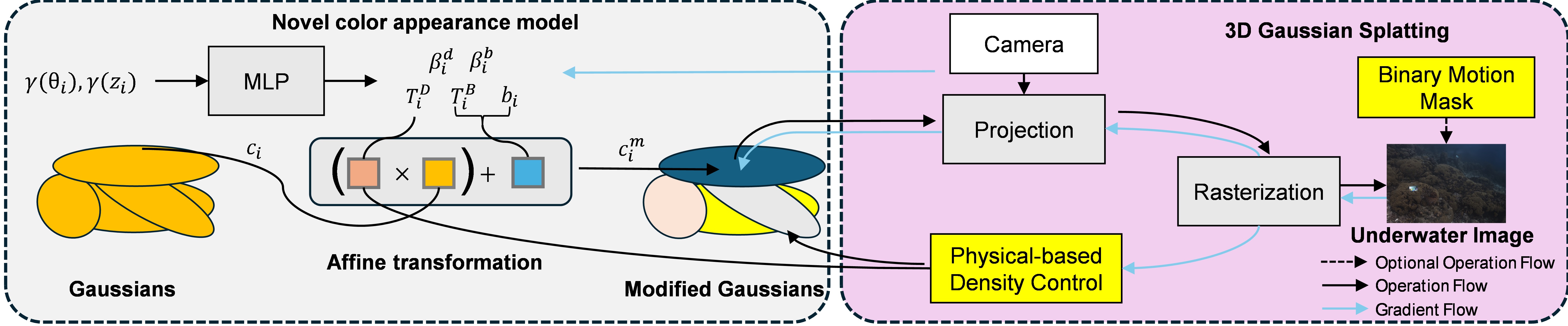}

   \caption{The diagram of our proposed UW-GS approach, combining a novel color appearance model, physical-based density control and binary motion mask to 3DGS. Our color appearance model uses view-direction $R$ and depth $z$ encoded by position encoding $\gamma$ to estimate water condition parameters: attenuation factor $T^{D}_{i}$, backscatter factors $T^{B}_{i}$, and medium coefficients $\beta^{d}_{i}$, $\beta^{b}_{i}$ and $b_{i}$. In the splatting process, the physical-based density control module addresses densification failures and the binary motion mask handle distractors.
   }
   \label{fig:workflow}
\end{figure*}

%------------------------------------------------------------------------
\section{Method}
\label{sec:formatting}

\subsection{Problem formulation}
3D Gaussian Splatting (3DGS) is a point-based approach that uses discrete Gaussian point clouds to represent a 3D scene. When rendering for a specific view, these 3D Gaussians are projected onto 2D space. Each Gaussian is modeled as follows: 
\begin{equation}
  G(x) = \exp(-\frac{1}{2}(x-\mu)^{\mathsf{T}}\Sigma^{-1}(x-\mu)),
  \label{eq:important}
\end{equation}
where $G$ is the Gaussian function with center position $\mu$ and 3D covariance matrix $\Sigma$. Each Gaussian is also characterized by opacity $\alpha$, view-dependent color $c$, expressed by spherical harmonics (SH). The $\alpha$-blending method is applied to generate the final image color, which is rasterized as described in \autoref{eq:originalmodel}:
\begin{equation}
  C = \sum_{i\in [1,N]} \alpha_{i}c_{i}\prod_{j=1}^{i-1}(1-\alpha_{j}),
  \label{eq:originalmodel}
\end{equation}
where $N$ refers to the number of Gaussians.

During training, the 2D position gradient, an intermediate result in backward propagation, is used as evidence to indicate “color under-represented” issues \cite{kim2024color}. Based on its average value, the sparse point cloud obtained from COLMAP \cite{schonberger2016structure} is adaptively grown by splitting and cloning.  Additionally, 3DGS control the number of Gaussians through periodic pruning to prevent over-densification. However, \autoref{eq:originalmodel} is insufficient for representing the underwater scenes \cite{tang2024neural}. Unlike clear air medium, the absorption and scattering effects in underwater environments can introduce additional attenuation and backscattering, impacting the scene representation. 
Following the suggestion in \cite{akkaynak2018revised}, underwater image intensity can be modeled as:
\begin{equation}
  I = J \cdot T^D + B^{\infty} \cdot (1 - T^B),
  \label{eq:revised}
\end{equation}
where $I$ represents the camera-captured color, $J$ is the true object color in the scene without medium effect, \textbf{$B^{\infty}$} denotes the background light color at the infinite distance. \textbf{$T^{D}$} and \textbf{$T^B$} reflect the impact of the medium (water) on the light coming directly from the object and the light that is backscattered by the water, respectively:
\begin{equation}
\begin{split}
  T^D &= \exp(-\beta^{d} \cdot z),\\
  T^B &= \exp(-\beta^{b} \cdot z),
\end{split}
  \label{eq:attnbs}
\end{equation}
where $\beta^{d}$ and $\beta^{d}$ are two RGB channel-wise medium coefficients, and $z$ is the distance. In conjunction with \autoref{eq:revised}, we infer that as an object moves further from the camera, its color (direct signal) fades while the background light (backscatter signal) increasingly dominates. Assuming the background light is ideally uniform, the color transition in areas far from the camera in underwater scenes is smooth, leading to color distortion and low contrast. As described in Mini-splatting \cite{fang2024mini}, this issue can undermine the effectiveness of adaptive density control used in the original 3DGS, which will be discussed in \autoref{sub:physicaldensity}. Consequently, only a limited number of sparse Gaussians are responsible for representing these regions, which often results in noticeable blur artifacts. Such issues are prevalent due to light absorption in underwater scenes. Therefore, we propose a new color appearance model and a physical-based density control module in UW-GS.

Moreover, moving objects such as fish and floating particles pose challenges to underwater 3D reconstruction. Their changing positions over time can cause errors in the appearance of static objects and introduce floaters in the rendering of novel views. Recent studies \cite{wu20244d, lin2024gaussian, yang2024deformable} have utilized temporal information to achieve 4D rendering; however, these methods typically model foreground motions with predictable motion trajectory and small displacements. In underwater environments, moving objects often exhibit fast speeds and significant positional changes, making dynamic rendering methods less suitable. Inspired by \cite{sabour2023robustnerf, zhou2024drivinggaussian}, we propose a pixel-level mask, named Binary Motion Mask (BMM), which detects outlier objects and eliminate them from the scene. 

\subsection{Overview of UW-GS}

The diagram of our UW-GS method is illustrated in \autoref{fig:workflow}, where our color appearance model is integrated into the 3DGS modified for underwater environments. The color appearance model using encoded depth and view direction as input for a multilayer perceptron (MLP) to estimate the parameters of the water medium and adjust the color of the Gaussian accordingly using an affine transformation. The 3D Gaussians with modified color will be sent to do 2D projection and then generate pixel color in rasterization module to output the final underwater image. 
The 3DGS includes a novel physics-based density control module using water medium estimation of MLP and a Binary Motion Mask (BMM) during training to mitigate the negative effects of the water medium and to handle moving objects, respectively.

\subsection{Color Appearance Model}

The original Gaussian color appearance in 3DGS \cite{kerbl20233d} is derived from SHs, which assumes the scene medium is transparent and struggles to model varying attenuation and backscattering effects in water. To address this issue, we propose a novel approach for color appearance formation. The left panel of \autoref{fig:workflow} illustrates the workflow of our method. Similar to \cite{lee2024compact}, we use an additional MLP $f$ with positon encoded depth and viewing direction input to estimate medium properties: 
\begin{equation}
  (T^{D}_{i}, T^{B}_{i}, \beta^{d}_{i}, \beta^{b}_{i}, b_{i}) = f(\gamma(z_{i}), \gamma(\theta_{i})), \quad i\in [1,N].
  \label{eq:MLP}
\end{equation}
Here $z_{i}$ and $\theta_{i}$ refer to the depth and view direction of Gaussian, respectively. We represent these by measuring the length and the normal vector of the line connecting the centers of the Gaussian and the camera. By using five head layers after the backbone, we can estimate the attenuation and backscatter factors $T^{D}_{i}$ and $T^{B}_{i}$, as well as medium coefficient per Gaussian: $\beta^{d}_{i}$, $\beta^{b}_{i}$ and $b_{i}$. These coefficients regulate two previous factors based on the physical model. Using the parameters obtained, the revised color $c^m$ is then calculated by applying an affine transformation as follows:
\begin{equation}
  c^{m}_i = T^{D}_i \cdot c_i + (1 - T^{B}_i) \cdot b_i,
  \label{eq:affinetransform}
\end{equation}

\noindent This model enables us to adjust object appearance to model visibility that varies with distance in underwater scene. The modified color for Gaussians is then sent to 3DGS module. 

\subsection{Physical-based Density Control}
\label{sub:physicaldensity}
\begin{figure*}[ht]
  \centering

   \includegraphics[width=1.0\linewidth]{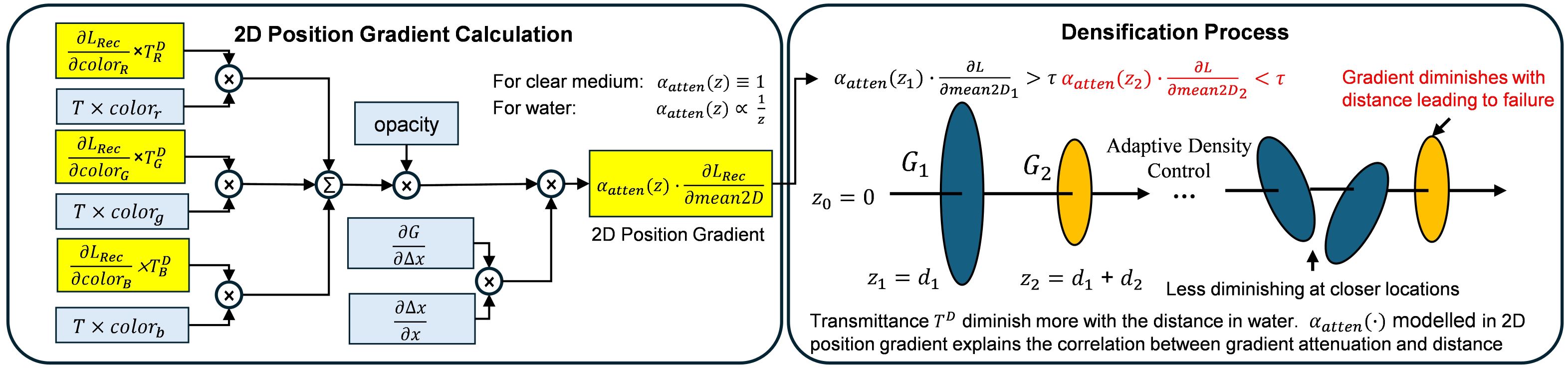}

   \caption{Left: Diagram of 2D Position gradient calculation. Right: Illustration of densification failures ($G_2$ highlighted in orange) that appear to be not cloned or split in underwater scenes. Please Note $color_{R,G,B}$ refers to color in pixel level while $color_{r,g,b}$ represents Gaussian color derived from SHs, which is considered as object color. $T$ refers to $\prod_{j=1}^{i-1}(1-\alpha_{j})$ in \autoref{eq:originalmodel} for $G_{i}$.}
   \label{fig:failfar}
\end{figure*}

In the standard 3DGS framework, the Gaussian point cloud is densified adaptively to acquire better representation capability. The 3DGS employs 2D position gradient $\frac{\delta L_{Rec}}{\delta \text{mean2D}}$, which is derived from pixel color gradients $\frac{\delta L_{Rec}}{\delta \text{color}}$. If this gradient exceeds a threshold $\tau$, the Gaussian is split or cloned. 
%In the standard 3DGS framework, the reconstruction loss $L_{Rec}$, which assesses rendering outcomes, combines L1 and D-SSIM losses. These losses produce channel-wise color gradients $\frac{\delta L_{Rec}}{\delta \text{color}}$ for each Gaussian. Typically, gradients such as the 2D position gradient $\frac{\delta L_{Rec}}{\delta \text{mean2D}}$ for density control are derived from this color gradient. 
However, in underwater scenes,  \autoref{eq:revised} suggests that $\frac{\delta L_{Rec}}{\delta color}$ is calculated from $T^{D} \cdot \frac{\delta L_{Rec}}{\delta color^{object}}$ so that each Gaussian color gradient has an attenuation factor $T^{D}$ due to light absorption effects. As shown in the left block in \autoref{fig:failfar}, the attenuation of the color gradient will result in the biased 2D position gradient calculation; we represent this effect as a function $\alpha_{atten}(z)$. Similar to $T^{D}$, this function is inverse correlated with depth $z$, meaning in underwater scenes, the 2D position gradient will diminish with distance. As a result, Gaussians can hardly be split or cloned in the far regions, leading to densification failures. This issue causes the blur artifact mentioned previously because of the sparse Gaussian distribution as shown in the right block of \autoref{fig:failfar}. 

%This however leads to poor reconstruction in area far from the camera, notably causing the blur artifact mentioned previously due to densification failures. As outlined in \cite{kim2024color} and shown in \autoref{fig:failfar}, the original 2D position gradient, used as a densification indicator, diminishes significantly with increased depth due to the attenuation factor in $\frac{\delta L_{Rec}}{\delta \text{color}_i}$, which leads to the biased 2D position gradient calculation result with an attenuation factor $\alpha_{atten}$. This issue result in failures of density control.

To mitigate the impact of the attenuation factor on gradient calculations, we introduce a refined densification strategy termed physics-based density control. For 2D position position gradient calculation, this approach employs the inverse of the estimated attenuation factor from novel appearance color model $ \frac{1}{T^{D}}$ to the underwater color gradient modeled as $T^{D} \cdot \frac{\delta L_{Rec}}{\delta color^{object}}$. Therefore, our $\frac{\delta L_{Rec}}{\delta \text{color}} = \frac{\delta L_{Rec}}{\delta color^{object}}$. This adjustment effectively compensates for biased 2D position gradient calculations, allowing 3DGS to generate denser Gaussian point clouds with improved representational accuracy. Furthermore, by adjusting the 2D position gradient, this method optimizes regions where object color information is substantially attenuated, thereby enhancing 3DGS performance in areas characterized by smooth color transitions. 

To further address the blur artifacts caused by large-size Gaussian projections, we follow \cite{zhang2024pixel} and consider the number of pixels covered in 2D space, rather than relying solely on view count to calculated average 2D position gradients for density control. This allows areas that were previously represented by sparse Gaussians in vanilla 3DGS to grow more effectively. As a result, the point clouds at larger scales exhibit improved representational capability, thereby reducing both blur and needle-like artifacts. 

Additionally, to deal with floaters close to the camera, which is a common issue of 3DGS, inspired by \cite{zhang2024pixel}, we topically reduce the 2D position gradients for Gaussians with a scaling factor $S_z$ related to depth $z$.  Hence, our final pixel color gradients are that $\frac{\delta L_{Rec}}{\delta \text{color}} = S_z \frac{\delta L_{Rec}}{\delta color^{object}}$. Combined with our physics-based density control, which scales gradients globally, this helps to suppress floaters near the camera without affecting our enhancement in other regions.

\subsection{Binary Motion Mask}
\label{sssec:BMM}
Although adopting the refined appearance model improves the modeling of underwater scenes, handling dynamic content in the dataset remains challenging. Some studies, such as \cite{azzarelli2023waveplanes, huang2024sc}, have modeled moving objects by incorporating an additional time dimension. However, moving underwater content, such as fish and flickers, typically exhibits random trajectories and high speeds. Dynamic scene rendering algorithms that use a deformation field network are neither efficient nor effective in these cases.  

Inspired by RobustNeRF \cite{sabour2023robustnerf}, we introduce a Binary Motion Mask (BMM) $\omega$  into our reconstruction loss function to eliminate the distractors as the follows:
\begin{equation}
  L_{Rec} = \lambda \cdot \omega \cdot L_{1} + (1-\lambda) \cdot \omega \cdot L_{D-SSIM},
  \label{eq:weighted3DGSRec}
\end{equation}
where $\lambda$ is a loss weight.

As indicated in \cite{sabour2023robustnerf}, the trimmed estimator is very effective in detecting a portion of the outliers in the set. In our implementation, $\omega_t$ at the iteration of $t$ is computed as follows:
\begin{equation}
    %\omega^{t}_1 = \epsilon^{t} \leq T_{\epsilon}, \quad T_{\epsilon} = \textrm{med}(\epsilon^{t-1}),\\
    \omega^{t}_1 = \epsilon^{t} \leq T_{\epsilon}, \\
  \label{eq:m1}
\end{equation}
where $\epsilon^{t}$ is a residual, computing from $\epsilon^{t} = \| \hat{I}_t-I\|_{2}$, and $\hat{I}_t$ is the rendered image at the iteration $t$.

In \autoref{eq:m1}, we hypothesize that residuals below the a threshold $T_{\epsilon}$, set by dividing sorted residuals from the previous iteration, belong to the inliers, while the rest are considered outliers. However, relying solely on this mask is insufficient to effectively distinguish distractors, which may introduce distortion in our rendering. For this reason, a $3\times3$ diffusion kernel $B$ is applied to maintain spatial smoothness as shown in \autoref{eq:m2}. Moreover, to avoid high-frequency content being treated as outliers, every $8\times8$ patch $R_{8\times8}$ is classified according to the average value of its $16\times16$ neighbors $R_{16\times16}$. The third mask is then described as shown in \autoref{eq:m3}.
\begin{equation}
    \omega^{t}_{2} = \omega^{t}_{1} \ast B \geq T_{\ast},
  \label{eq:m2}
\end{equation}
\begin{equation}
    \omega^{t}_{3}(R_{8\times8})  = \overline{\omega^{t}_{2}(R_{16\times16})} \geq T_{R},
  \label{eq:m3}
\end{equation}
where $T_{\ast}$ and $T_{R}$ are thresholds and can be manually tuned like $T_{\epsilon}$. If a pixel is classified as an inlier by any of the three weighting formulas described above, it is ultimately regarded as an inlier to prevent our estimation from being overly aggressive.
The final BMM at $t$ is then obtained from
\begin{equation}
    \omega^t  = \omega^t_1 \cup \omega^t_2 \cup \omega^t_3.
  \label{eq:omega}
\end{equation}

\subsection{Loss Function}
\label{ssec:lossfunction}
The final loss function, $\mathcal{L}$, used for optimization, comprises a reconstruction loss $L_{Rec}$ (see \autoref{sssec:BMM}), a depth-supervised loss $L_{da}$, and a gray-world assumption loss $L_{g}$, defined as follows:
\begin{equation}
    \mathcal{L} = L_{Rec} + L_{da} + L_g.
    \label{eq:final_loss}
\end{equation}

\noindent\textbf{Depth-supervised loss $L_{da}$.} Depth information is the key to underwater scene modeling. In 3DGS, we can render depth maps in a similar way as image:
\begin{equation}
  \hat{D} = \sum_{i\in N} \alpha_{i}z_{i}\prod_{j=1}^{i-1}(1-\alpha_{j}).
  \label{eq:rendereddepth}
\end{equation}
However, unreasonable positioned Gaussian sets, such as floaters, will heavily affect our estimation of the underwater medium coefficients due to the erroneous depth measurements. In order to enhance the reliability of the depth measurements between the Gaussians mean center and the camera center and diminish the floater issue. We implement a depth-supervised loss, denoted as $L_{da}$:
\begin{equation}
\begin{split}
        L_{da} &= \lambda_{d}L_{d} + \lambda_{ca}L_{ca}, \\
        L_{d} = \| \hat{D} - D\|, \: \: L_{ca} &=\sum_{x\in\{r,g,b\}} \sum_{y\in\{d,b\}} \|\hat{z}_{x}^{y} - z\|,\\
\end{split}
  \label{eq:depthsupervision}
\end{equation}
where $ \lambda_{d}$ and $\lambda_{ca}$ are hyperparameters, and $L_{d}$ is a depth supervison L1 loss. We use a synthetic ground truth depth map $D$ that is predicted by DepthAnything \cite{yang2024depth}, a novel monocular depth estimation model that can adapt to any environment. Please note Both $D$ and $\hat{D}$ are normalized to aligned relatively instead of absolutely. Inspired by \cite{varghese2023self}, $L_{ca}$, named channel-wise depth alignment loss is also included. This loss is developed based on the underwater image formation model~\autoref{eq:revised}, resulting in the approximate depth per Gaussian can be calculated in the other way: $\hat{z}_{x}^{y} = -\frac{\log(T_{x}^{y})}{\beta^{y}}$. $z$ is the measured depth of the Gaussian. $L_{ca}$ imposes restrictions on $T^D$ and $T^B$, which are closely related to the depth. Here in $L_{da}$, we correct depth information at both the pixel and Gaussian point levels. 

\noindent\textbf{Gray-world assumption loss $L_{g}$.} A gray-world assumption loss $L_{g}$ based on the image restoration prior knowledge \cite{fu2022unsupervised} is also incorporated:
\begin{equation}
    L_{g} = \sum_{x\in(r,g,b)}\| \mu(J^{x})- 0.5\|_{2}^{2}.
  \label{eq:grayworld}
\end{equation}
Grounded in the Gray-world assumption, $L_{g}$ serves to constrain the color of objects within the scene that have similar average values across the RGB channels, with the objective of ensuring that $\beta^{d}$, $\beta^{b}$ and $b$ are predicted in an appropriate manner.

\section{Experiment Configuration}
%This section specifies the experiment designed to evaluate the proposed method, including databases, implementation details, and baseline methods.

\begin{figure}[!t]
  \centering
   \includegraphics[width=1.0\linewidth]{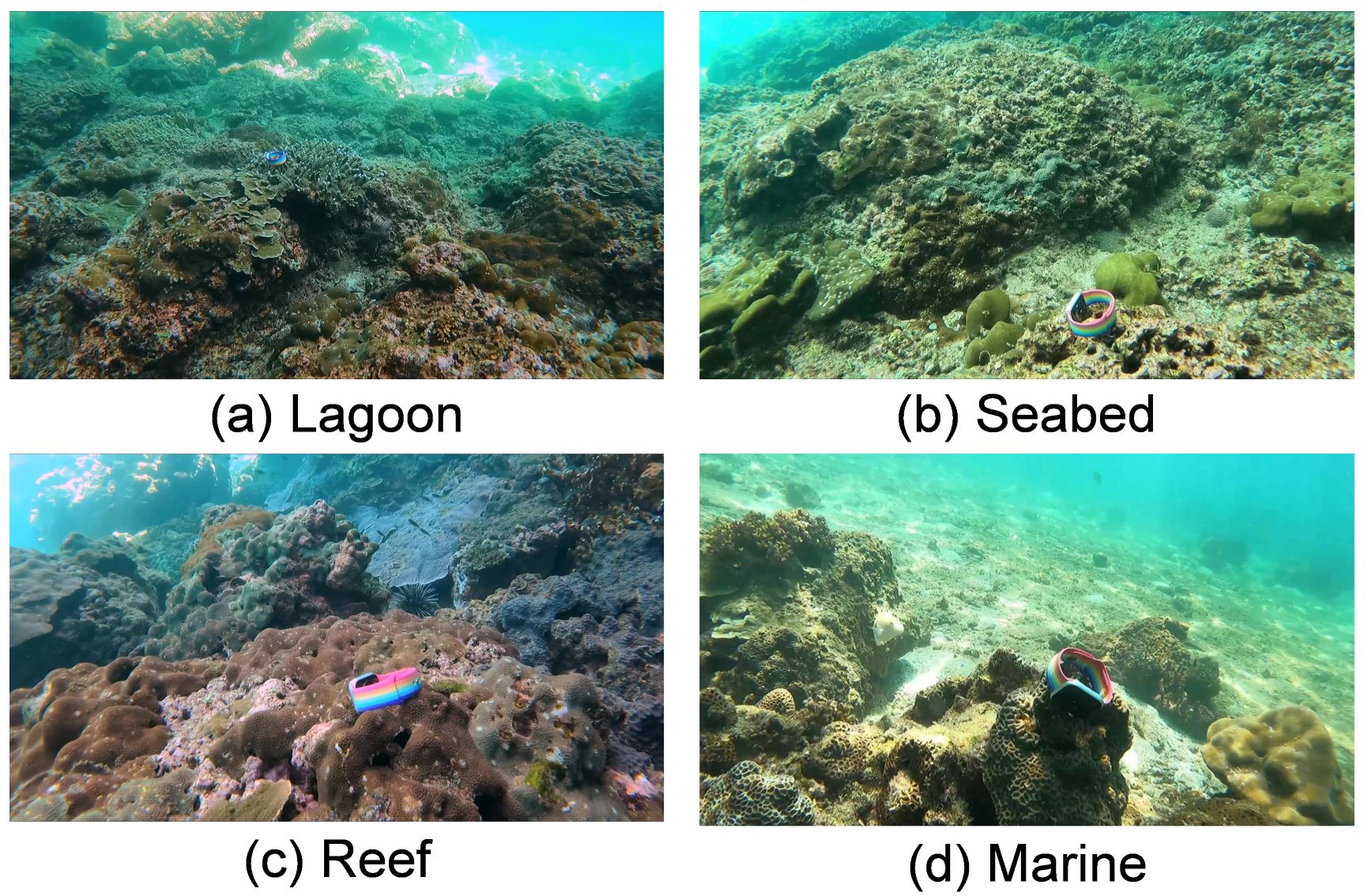}
   \caption{Sample images from S-UW dataset.}
   \label{fig:S-uw}
\end{figure}

\noindent\textbf{Datasets.} Three datasets were used for performance evaluation: i) SeaThru-NeRF \cite{levy2023seathru}; ii) the in-the-wild (IW) dataset from \cite{tang2024neural}, and iii) our dataset, S-UW.
SeaThru-NeRF \cite{levy2023seathru} comprises four image sequences captured in different seas, each pre-processed with white balance enhancement to reduce noise. The IW dataset \cite{tang2024neural} was captured at mid-depth underwater, with moving object masks available. This dataset was used for evaluating the BMM performance.
 
Our S-UW dataset consists of high-definition underwater videos across four larger areas scenarios, each containing 24 images. Examples of these scenes are shown in \autoref{fig:S-uw}. Unlike existing datasets, S-UW was captured in shallow waters and introduces additional challenges, such as flickering. 
 %We sorted the sequence and 
 For all videos, we selected one image out of every eight images for testing.

\noindent\textbf{Implementation Details.} Our method is implemented using Pytorch. We utilized COLMAP \cite{schonberger2016structure} to initialize the point cloud and estimate the camera position of the images in the sequence. We modified the diff-gaussian-rasterization used in original 3DGS to render depth maps, incorporating the additional gradient calculation branch in a backward process for depth loss. 
For the medium MLP, we use two linear layers with 64 features and additional head layers for each output. Softplus activation is adopted for $\beta^{D}$ and $\beta^{B}$ while Sigmoid is used for $T^D$, $T^B$ and $b$.

For training, we carried out 15,000 iterations in a single RTX3090 GPU. The first 1,000 iterations are used as a warm-up phase, in which we optimize the model by solely using depth loss $L_{d}$ and depth-supervised loss $L_{da}$ to adjust the initialized point cloud and pre-train MLP. Subsequently, we used the same optimization setting as the original 3DGS to optimize Gaussian attributes and applied an independent Adam optimizer to train our medium MLP. The BMM, is applied to the scenes containing dynamic objects.

\begin{table*}[!t]
\centering
\small  %
\resizebox{1.0\linewidth}{!}{
\begin{tabular}{r|ccc|ccc|ccc|ccc}
		\midrule
              \multicolumn{13}{c}{SeaThru-NeRF}\\
\toprule
Scene      & \multicolumn{3}{c|}{Curasao}& \multicolumn{3}{c|}{Panama}& \multicolumn{3}{c|}{IUI-Reasea}& \multicolumn{3}{c}{Japanese-Reasea} \\
Method & PSNR $\uparrow$ & SSIM $\uparrow$ & LPIPS $\downarrow$ & PSNR $\uparrow$ & SSIM $\uparrow$ & LPIPS $\downarrow$& PSNR $\uparrow$ & SSIM  $\uparrow$& LPIPS $\downarrow$ & PSNR $\uparrow$ & SSIM $\uparrow$ & LPIPS $\downarrow$ \\
\hline
Instant-NGP \cite{muller2022instant} & 27.9051 & 0.7069  & 0.3853 & 23.4593  & 0.6359  & 0.4262 & 20.6342  & 0.5577  & 0.6027& \textbf{23.2302}  &  0.6551  & 0.3567  \\

SeaThru-NeRF \cite{levy2023seathru}& 29.9210  & 0.8563  & 0.2976 & 26.9008  & 0.7893  & 0.3297 & \underline{25.8899}  & 0.7537  & 0.3534 & 21.7473  & 0.7351  & 0.3367  \\
3DGS \cite{kerbl20233d}& \underline{30.9733} & \underline{0.9356}  & \underline{0.1930} & \underline{30.8030}  & \underline{0.9186}  & \underline{0.1963} & 23.1343  & \underline{0.8774}  & \underline{0.2402} & 1.9711  & \textbf{0.8671}  & \underline{0.2018}  \\
UW-GS (ours) & \textbf{31.7714}  & \textbf{0.9430}  & \textbf{0.1441} & \textbf{31.7902}  & \textbf{0.9360}  & \textbf{0.1156} & \textbf{28.6523}  & \textbf{0.9334}  & \textbf{0.1252} & \underline{23.0465}  &  \underline{0.8602}  & \textbf{0.1899}  \\
            \midrule
            \multicolumn{13}{c}{S-UW}\\\midrule
            
Scene      & \multicolumn{3}{c|}{Lagoon}& \multicolumn{3}{c|}{Seabed}& \multicolumn{3}{c|}{Reef}& \multicolumn{3}{c}{Marine} \\
Method & PSNR $\uparrow$ & SSIM $\uparrow$ & LPIPS $\downarrow$ & PSNR $\uparrow$ & SSIM $\uparrow$ & LPIPS $\downarrow$& PSNR $\uparrow$ & SSIM  $\uparrow$& LPIPS $\downarrow$ & PSNR $\uparrow$ & SSIM $\uparrow$ & LPIPS $\downarrow$ \\
\hline
Instant-NGP \cite{muller2022instant}& 21.6674  & 0.5480  & 0.3198 & 19.6167  & 0.4275  & 0.3198 & 18.5866  & 0.4382  & 0.5058 & 15.7300  & 0.4637  & 0.4410  \\

SeaThru-NeRF \cite{levy2023seathru} & 25.5424  & 0.7592  & 0.2860 & 22.7050  & 0.6407  & 0.3455 & 18.5510  & 0.4077  &  0.5522 & 18.1031  & 0.5718  & 0.4582   \\
3DGS \cite{kerbl20233d}&\textbf{27.1166}  & \underline{0.8707}  & \underline{0.1606} & \underline{26.4933}  & \textbf{0.8234}  & \underline{0.1684} &  \underline{22.4663}  & \underline{0.7117}  & \underline{0.2646} & \underline{18.9492}  & \underline{0.6922}  & \underline{0.2990}  \\
UW-GS (ours) & \underline{26.9103}  & \textbf{0.8736}  & \textbf{0.1479} & \textbf{26.5088}  & \underline{0.8197}  & \textbf{0.1541} & \textbf{23.3078}  & \textbf{0.7277}  & \textbf{0.2504} & \textbf{20.0065}  & \textbf{0.7145}  & \textbf{0.2678} 
\\\bottomrule
\end{tabular}}
\caption{Qualitative results of the proposed method evaluated on SeaThru-NeRF and S-UW dataset. $\uparrow$ refers larger values are better while $\downarrow$ is opposite. \textbf{Bold} indicates the best results and \underline{underline} values represent the second best.}
\label{tab:seathruNeRFdataset}
\end{table*}

\begin{figure*}[!t]
  \centering

   \includegraphics[width=1.0\linewidth]{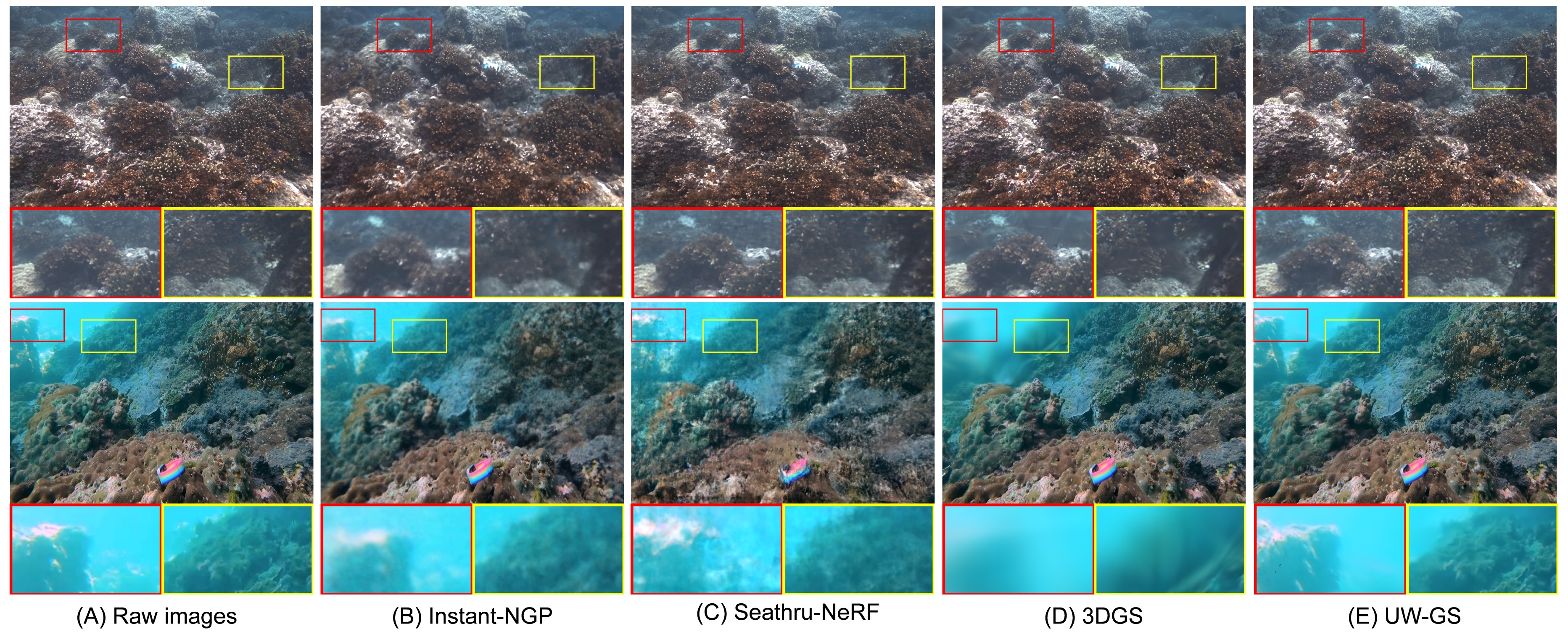}
 %\missingfigure[figwidth=\linewidth,figheight=0.3\linewidth]{resulttwolines}
 
       \caption{Novel view rendering comparison in \textbf{Panama} from Seathru-NeRF dataset \cite{levy2023seathru} and \textbf{Reef} from S-UW. We have shown details blow the images. Please note that images in first row are enhanced for visualization.}
   \label{fig:Result}
\end{figure*}

\noindent\textbf{Baselines and Metrics.} We tested our method and compared with three state of the arts:  Instant-NGP \cite{muller2022instant}, SeaThru-NeRF\cite{levy2023seathru}, and original 3DGS \cite{kerbl20233d}. We used their official implementation, but trained on the same sequence using the same dataset split strategy. 
%To validate the effectiveness of our binary motion mask, we will only test it on the dataset from \cite{tang2024neural}. Using the motion masks provided in the dataset as the ground truth, 
We computed the global PSNR, SSIM and LPIPS values in the static areas. On the other hand, we will also use these three metrics in dynamic scenes after using motion mask provided from dataset to exclude moving objects.

\section{Results and Discussion}
\label{ssec:results}

%This section provides the quantitative and qualitative results obtained in the experiment and performs and ablation study to verify the major contributions of this work.

\noindent\textbf{Quantitative results.}
%First, to validate the performance of our method, we tested it on both SeaThru-NeRF and proposed S-UW datasets. 
\autoref{tab:seathruNeRFdataset} shows the evaluation results of the SeaThru-NeRF and S-UW datasets for the rendered quality of each benchmark. For the SeaThru-NeRF dataset, our method shows the best overall performance and achieves average 2.09dB and 2.70dB PSNR improvement compared to 3DGS and Seathru-NeRF respectively, although it has the second-best SSIM in the Japanese-RedSea scene. 
%\hl{Add why our results are the best and why SSIM slighly less in Japanese scene. Calculate the average PSNR (no need to add in the table) and say how much in \% UW-GS is bettern than 3DGS and SeaThru-NeRF} 
For the S-UW dataset, while UW-GS ranks second in PSNR for the Lagoon and in SSIM for the Seabed scene, it still performs best and exceeds the average PSNR of 3DGS and Seathru-NeRF by 0.43 dB and 2.96 dB, respectively. The limited improvement compared to 3DGS can be attributed to the unstable lighting from above the water surface.
%\hl{HW: describe the results in Table 1.}

\noindent\textbf{Qualitative results.}
\autoref{fig:Result} presents a visual comparison between the baseline methods and UW-GS. The first row shows the results for the SeaThru-NeRF dataset, while the second row displays the results for the S-UW dataset. In summary, our method enhances the quality of novel view synthesis results with better reconstruction of distant regions and fewer artifacts. Our approach illustrates effectiveness and robustness across various scenes.

\begin{figure}[!t]
  \centering
   \includegraphics[width=0.9\linewidth]{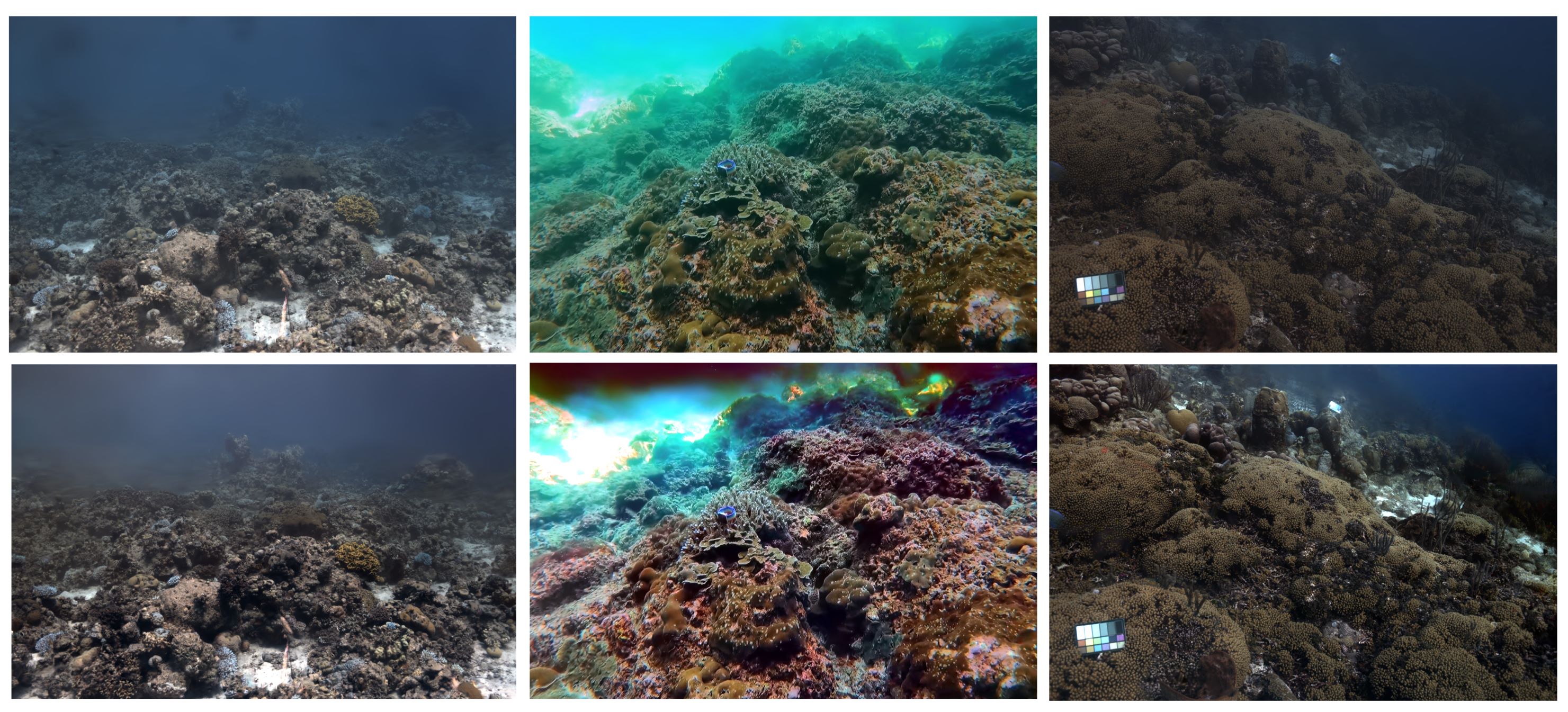}
   \caption{The first row presents render images while the second row shows their corresponding estimated clean images $\hat{J}$, which are obtained by solely using the color of Gaussian in rasterization without transformation. From left to right come from \textbf{Japanese-Redsea}, \textbf{Lagoon} and \textbf{Curasao} scene respectively.}
   \label{fig:estimatedcleanimage}
\end{figure}

In addition, our method demonstrates its capability to distinguish objects from the scattering medium. In this context, we attempted to “drain” the water to present clearer restored results based on the estimated water medium acquired from our model. \autoref{fig:estimatedcleanimage} shows that our method has potential in the enhancement of underwater scenes.

%Furthermore, to verify our method of moving object removal, we tested it on two challenging scenes from the dataset proposed by \cite{tang2024neural}, which contain multiple fish of varying sizes. In this experiment, we additionally applied the binary motion mask to our reconstruction loss to estimate motion during the training. In \autoref{fig:BMM}, we compare the rendered results of using and without a BMM in \autoref{fig:compareBMM}; this mask is also visualized. In \autoref{fig:compareBMM}, we assess the reconstruction performance excluding moving objects by using ground motion masks from the dataset. To conclude, our method has proven its outstanding performance in removing the dynamics from the scene rendering without adopting any additional neural networks. However, because BMM cannot make classification perfectly, the reconstruction quality of static areas will be degraded to some degree.

\begin{figure}[!t]
  \centering

   \includegraphics[width=0.9\linewidth]{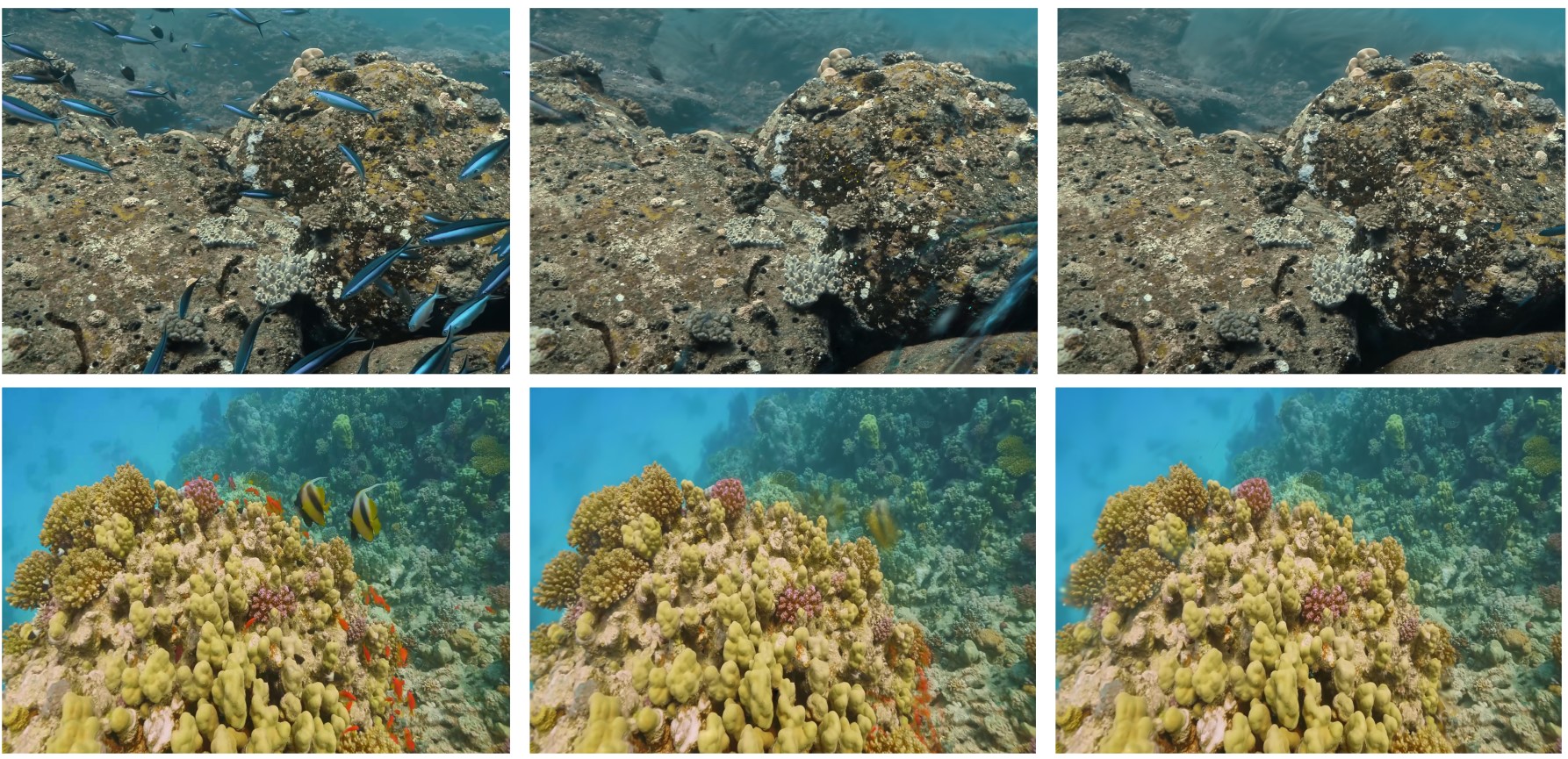}
 %\missingfigure[figwidth=\linewidth,figheight=\linewidth]{BMM results}
 
   \caption{Examples of rendering results from \textbf{Composite} and \textbf{Sardine} scenes. From left to right: raw videos, results without and with BMM, respectively. }
   \label{fig:compareBMM}
\end{figure}

\noindent\textbf{Ablation study.}
We isolate our contributions using a set of modified architectures: (V1) solely using spherical harmonics to represent view-dependent color (note that MLP will also assist with density control), (V2) training UW-GS without physics-based density control, and (V3) removing the depth-supervised loss. The results are shown in \autoref{tab:ablationstudy1} verifies that our architecture can achieve the overall best performance while others suffer from performance degradation.%\hl{Need discussion here too.}

We also analyzed how the three masks used in the BMM contribute to moving object removal. We selected two challenging scenes from the IW dataset \cite{tang2024neural}, which includes motion masks that aid in assessing reconstruction quality by excluding dynamic content. \autoref{tab:BMMmask} indicates our method yields the best performance. Demonstrate BMM combining three masks can best retain static content during training.

\begin{table}[!t]
\small
    \centering
    \begin{tabular}{r|ccc}
    \midrule
        Configuration & PSNR$\uparrow$& SSIM$\uparrow$& LPIPS$\downarrow$\\
        \hline
        V1 & 25.4132 & 0.8338 & \underline{0.1884} \\
        V2 & \underline{25.9165} & 0.8313 & 0.2115 \\
        V3 & 25.7046 & \underline{0.8392} & 0.2026 \\
        Ours & \textbf{26.4982} & \textbf{0.8510} & \textbf{0.1744} \\\bottomrule
    \end{tabular}
    \caption{Ablation Study on SeaThru-NeRF and S-UW Scenes using our proposed methods and its modified architectures}
    \label{tab:ablationstudy1}
\end{table}

\noindent\textbf{Limitations.}
The improvement of our method is not obvious in the shallow underwater scene because the disturbance of light from above the water cannot be neglected. % In additon, the coefficients of BMM calculation are adjusted manually. In the next step, we may need to make those parameters trainable. We still suffers from training and rendering speed degradation. In the future, we need to focus on simplification.
Subjectively, our method eliminates dynamic distractors from scene renderings without any additional pre- or post-processing. More subjective results are included in supplementary materials. However, because BMM may imperfectly classify, the reconstruction quality in static areas is somewhat degraded, as seen in the objective results of IW dataset shown in \autoref{tab:compareBMM}.

\begin{table}[!t]
\small
    \centering
    \begin{tabular}{l|ccc}
            \toprule
        \multicolumn{4}{c}{BMM Break Down}\\\midrule
        Configuration & PSNR$\uparrow$& SSIM $\downarrow$ & LPIPS$\downarrow$\\
        \hline
        + $\omega_{1}$ & 24.1491 & 0.8568 & 0.1625\\
        + $\omega_{2}$ & \underline{24.6537} & \underline{0.8943} & \underline{0.1251}\\
        + $\omega_{3}$ (Ours) & \textbf{25.9267} & \textbf{0.9081} & \textbf{0.1098} \\
        \bottomrule
     \end{tabular}
    \caption{Performance on different marks in BMM, training in \textbf{Composite} and \textbf{Sardine} scenes \cite{tang2024neural}. Motion masks are provided with the dataset.}
    \label{tab:BMMmask}
\end{table}

\begin{table}[!t]
\small
    \centering
    \begin{tabular}{l|ccc}
        \toprule
        method & PSNR & SSIM & LPIPS \\
        \midrule
        UW-GS without BMM & 27.8302 & 0.9231 & 0.0960 \\
        UW-GS using BMM & 25.9267 & 0.9081 & 0.1098 \\
        \bottomrule
    \end{tabular}
    \caption{The average results of \textbf{Composite} and \textbf{Sardine} scenes \cite{tang2024neural} using and without a BMM during training.}
    \label{tab:compareBMM}
\end{table}

%\begin{table}[!t]
%\small
%    \centering
%    \begin{tabular}{l|ccc}
% 		\toprule
%              \multicolumn{4}{c}{Composite}\\
%\hline
%        Configuration & PSNR$\uparrow$& SSIM$\uparrow$& LPIPS$\downarrow$ \\
%        \hline
%        UW-GS without BMM & 28.4158 & 0.9173 & 0.0957 \\
%        UW-GS using BMM & 25.9188 & 0.8916 &  0.1168\\
%                    \midrule
%            \multicolumn{4}{c}{Sardine}\\\midrule
%        Configuration & PSNR$\uparrow$& SSIM$\uparrow$& LPIPS$\downarrow$ \\
%        \hline
%        UW-GS without BMM & 27.2447 & 0.9290 &  0.0964 \\
%        UW-GS using BMM & 25.9346 & 0.9246 & 0.1028 \\
%   \bottomrule
%    \end{tabular}
%    \caption{The average results of using and without a BMM during training in \textbf{Composite} and \textbf{Sardine} scenes \cite{tang2024neural}.}      
%    \label{tab:compareBMM}
%\end{table}

%\begin{figure}[!t]
%  \centering
%   \includegraphics[width=1.0\linewidth]{figure/mask.jpg}
%   \caption{Examples from BMM results during training.}
%   \vspace{-0.5em}
%   \label{fig:BMM}
%\end{figure}

%\begin{table}[!t]
%\small
    %\centering
    %\caption{Ablation Study on dataset from \cite{tang2024neural} using our proposed methods. }
    %\label{tab:ablationstudy2}
    %\begin{tabular}{l|ccc}
    %\midrule
        %Configuration & PSNR$\uparrow$& SSIM $\downarrow$ & LPIPS$\downarrow$\\
        %\hline
        %+ $\omega_{1}$ & 24.1491 & 0.8568 & 0.1625\\
        %+ $\omega_{2}$ & \underline{24.6537} & \underline{0.8943} & \underline{0.1251}\\
        %+ $\omega_{3}$ (Ours) & \textbf{25.9267} & \textbf{0.9081} & \textbf{0.1098} \\\bottomrule
    %\end{tabular}
%\end{table}

\section{Conclusion}

This paper presents a novel 3DGS-based method for underwater applications. We introduce a novel color appearance model to better represent distance-dependent attenuation and backscatter, which significantly enhances reconstruction results, especially in far regions strongly influenced by water effects. Additionally, we propose the BMM that effectively eliminate the shadow-like artifacts due to the distractors by excluding them during the training. We also introduce a challenging underwater dataset, comprising 4 different scenes. Our method is tested on several scenes from three datasets and outperforms the existing static underwater scene reconstruction baselines while demonstrating the ability to handle dynamics, highlighting its promising potential for underwater environment exploration and research. 

\vspace{5pt}

\noindent\textbf{Acknowledgments.} This work was supported by the UKRI MyWorld Strength in Places Program (SIPF00006/1) and the EPSRC ECR International Collaboration Grants (EP/Y002490/1).

%%%%%%%%% REFERENCES
{\small
\bibliographystyle{ieee_fullname}
\bibliography{PaperForReview}
}

\end{document}